\documentclass[twocolumn]{autart}    

\usepackage{mathtools}
\usepackage[hidelinks]{hyperref}
\usepackage{algorithmic}
\usepackage{algorithm}
\usepackage{amsmath,amssymb}
\usepackage{booktabs}
\usepackage{color}
\usepackage{cite}
  
\allowdisplaybreaks[4]

\newtheorem{lemma}{Lemma}
\newtheorem{theorem}{Theorem}
\newtheorem{assumption}{Assumption}

\newtheorem{proposition}{Proposition}

\newcommand{\EE}{\mathbb{E}}

\newcommand{\BP}{\mathbb{P}}

\newcommand{\PP}{\mathbb{P}}

\begin{document}

\begin{frontmatter}

\title{Distributionally Robust Federated Learning with \\ Outlier Resilience} 

\thanks[footnoteinfo]{This work was supported in part by Swedish Research Council Distinguished Professor Grant 2017-01078, Knut and Alice Wallenberg Foundation, Wallenberg Scholar Grant, Swedish Strategic Research Foundation SUCCESS Grant FUS21-0026, and AFOSR under award \#FA9550-19-1-0169. 
}

\author[DCS]{Zifan Wang}\ead{zifanw@kth.se},    
\author[TONGJI]{Xinlei Yi}\ead{xinleiyi@tongji.edu.cn},               
\author[DUKE]{Xenia Konti}\ead{polyxeni.konti@duke.edu},
\author[DUKE]{Michael M. Zavlanos}\ead{michael.zavlanos@duke.edu},
\author[DCS]{Karl H. Johansson}\ead{kallej@kth.se}

\address[DCS]{Division of Decision and Control Systems, School of Electrical Engineering and Computer Science, KTH Royal Institute of Technology,  Stockholm, Sweden}    
\address[TONGJI]{Department of Control Science and Engineering, College of Electronics and Information Engineering, Tongji University, Shanghai, China}    
\address[DUKE]{Mechanical Engineering and Material Science, Duke University, Durham, USA}       


\begin{abstract}                          
Federated learning (FL) enables collaborative model training without direct data sharing, but its performance can degrade significantly in the presence of data distribution perturbations. Distributionally robust optimization (DRO) provides a principled framework for handling this by optimizing performance against the worst-case distributions within a prescribed ambiguity set. However, existing DRO-based FL methods often overlook the detrimental impact of outliers in local datasets, which can disproportionately bias the learned models. In this work, we study distributionally robust federated learning with explicit outlier resilience. We introduce a novel ambiguity set based on the unbalanced Wasserstein distance, which jointly captures geometric distributional shifts and incorporates a non-geometric Kullback--Leibler penalization to mitigate the influence of outliers. This formulation naturally leads to a challenging min--max--max optimization problem. To enable decentralized training, we reformulate the problem as a tractable Lagrangian penalty optimization, which admits robustness certificates. Building on this reformulation, we propose the distributionally outlier-robust federated learning algorithm and establish its convergence guarantees. Extensive experiments on both synthetic and real-world datasets demonstrate the effectiveness of our approach.   

\end{abstract}

\end{frontmatter}

\section{Introduction}

Federated learning (FL) has emerged as a promising paradigm for collaborative model training in privacy-sensitive domains, such as mobile devices, healthcare networks, and industrial IoT systems \cite{mcmahan2017communication,kairouz2021advances,li2020federated}.
In FL, multiple clients compute model updates locally on private data, while a central server aggregates these updates to train a shared model without exchanging raw data.
However, client-specific variations typically result in heterogeneous data distributions.
Moreover, these local data distributions often experience covariant shifts over time \cite{duchi2019distributionally}.
This data heterogeneity, together with distribution shifts, poses significant challenges to FL, particularly for robustness and model generalization to unseen distributions.

To enhance the generalization performance, researchers have turned to distributionally robust optimization (DRO) \cite{delage2010distributionally,kuhn2025distributionally,levy2020large}.
The core idea of DRO is to optimize model performance against the worst-case distribution within a prescribed set of distributions, known as the ambiguity set.
The construction of this ambiguity set is crucial in the development of DRO methods, as it determines the type of distribution perturbations the model is robust to.
A common choice is to define the ambiguity set using the Kullback–Leibler (KL) divergence due to its computational benefits \cite{hu2013kullback,duchi2021learning}.
Since the KL divergence does not consider the geometry of a distribution’s support, KL-based DRO methods are effective for modeling probability reweighting but are less suited for capturing shifts with geometric features.
An alternative is to construct the ambiguity set using the Wasserstein distance, which naturally captures geometric shifts in the feature space.
The resulting Wasserstein DRO problem has been extensively studied and is supported by strong finite-sample guarantees under mild assumptions \cite{gao2023distributionally,mohajerin2018data,blanchet2019quantifying}.
It has been successfully applied to a variety of machine learning problems, including linear regression \cite{chen2018robust} and logistic regression \cite{shafieezadeh2015distributionally}.

In many practical scenarios, the observed data may contain outliers arising from adversarial attacks, sensor failures, labeling errors, or rare but legitimate events. Classical DRO methods are not suited for handling this issue, as even a small fraction of outliers can greatly distort the constructed ambiguity set and impede robust decision-making \cite{hughes2021increasing}. In FL, where datasets remain decentralized and private, outliers are especially problematic. Due to client-specific heterogeneity, data points considered normal for one client may act as outliers for others \cite{doi:10.1161/CIR.0000000000000628}, depending on the task context. For example, in a federated medical diagnosis task, physiological measurements from healthy young adults 
may be normal for an athletic cohort but flagged as outliers for a client representing elderly patients. This task-dependent variability makes outlier removal ineffective, as samples discarded by one client could represent valuable rare cases for another.

In this paper, we introduce distributionally outlier-robust federated learning (DOR-FL), the first FL framework to jointly address data heterogeneity, distribution shifts, and outlier contamination within a unified DRO formulation \cite{huber2011robust,ljung1998system}.
Our approach utilizes the unbalanced Wasserstein (UW) \cite{wangoutlier} distance, which replaces the hard constraints with soft penalization, thereby reducing the influence of outliers. Using the UW distance, we define a novel ambiguity set that captures both cross-client distribution shifts and robustness to outliers. The resulting DRO problem has a challenging min–max–max structure. To make it computationally tractable, we introduce a Lagrangian penalty reformulation that enables decentralized training. Moreover, we show that this reformulation preserves robustness supported by a robustness certificate. Then, we design a federated averaging–based algorithm with provable convergence guarantees under standard assumptions. Experiments on both synthetic and real-world datasets confirm the effectiveness of DOR-FL, showing substantial improvements over existing baselines.

Most closely related to our study is distributionally robust FL \cite{mohri2019agnostic,deng2020distributionally,yu2023scaff,zhang2023stochastic,nguyen2022generalization,jiao2023federated,konti2024distributionally,konti2025distributionally}. For example, \cite{mohri2019agnostic} introduced agnostic federated learning (AFL), which formulates FL as a minimax optimization problem to ensure good performance across all possible mixtures of client distributions. 
\cite{nguyen2022generalization} instead incorporates a Wasserstein-based ambiguity set to enable robustness to geometric perturbations centered at a fixed mixture of client distributions. 
\cite{konti2025distributionally} further proposes a unified framework that addresses both arbitrary mixtures of client distributions and geometric perturbations.
While these methods address geometric distribution shifts and client heterogeneity, they assume that local datasets are clean and therefore remain vulnerable to outlier contamination. In the presence of extreme or corrupted samples, the ambiguity sets in these formulations can be heavily distorted, leading to overly conservative models or degraded accuracy. Consequently, these approaches cannot be applied here.

This paper is also related to Byzantine-robust FL \cite{blanchard2017machine,zhu2023byzantine}, which defends against adversarial clients by treating entire client updates as potential outliers. This setting differs fundamentally from ours, as we address sample-level outliers within each client’s dataset rather than client-level corruption. Another relevant line of research is reweighted FL \cite{li2023revisiting,xu2023aggregation}, which adjusts the aggregation weight of each client to improve robustness against heterogeneous data quality. In contrast, our approach performs sample-level reweighting to directly downweight suspected data. Perhaps most closely related is \cite{wangoutlier}, which studies outlier-robust optimization in the centralized setting. Compared to \cite{wangoutlier}, we provide formal robustness certificates and, more importantly, extend the formulation to the federated setting. This extension transforms the optimization problem from a min–max structure to a more challenging min–max–max structure, requiring new algorithmic design and analysis to enable efficient decentralized training.

The remainder of this paper is organized as follows. In Section~\ref{sec:2_problem}, we introduce our outlier-robust federated learning framework. Section~\ref{sec:3_result} presents a tractable Lagrangian penalty reformulation of the proposed problem. In Section~\ref{sec:4_alg}, we develop the learning algorithm and establish its convergence guarantees. Section~\ref{sec:5_experiments} reports experimental results on both synthetic and real-world datasets to validate the effectiveness of our approach. Finally, Section~\ref{sec:6_conclusion} concludes the paper.
\section{Problem Statement}
\label{sec:2_problem}

\subsection{Distributionally Robust Optimization}
Consider the stochastic optimization problem 
\begin{align*}
    \min_{\theta \in \Theta} \EE_{ \PP^*} [l(\theta,\xi)]
\end{align*}
where the random samples $\xi \in \Xi$ are drawn from a fixed distribution $\BP^* \in \mathcal{P}(\Xi)$, $\mathcal{P}(\Xi)$ is the set of distributions on the support $\Xi$, $\theta \in \Theta$ is the decision variable, and $l$ is the loss function. 
In many applications, the true distribution of interest, $\BP^*$, is not explicitly known. A dataset comprising finite samples independently drawn from $\BP^*$ are often available,  which yield an empirical distribution estimate $\hat{\BP}$.
However, the empirical data distribution $\hat{\PP}$ often fail to reliably approximate the true distribution $\PP^*$, especially in the presence of distribution shifts.

To address distribution shifts, Wasserstein DRO optimizes the worst-case performance over all distributions within an ambiguity set, which is constructed around the empirical distribution. The ambiguity set aims to include highly probable distributions $\PP^*$ to ensure robustness while excluding less probable distributions to reduce conservatism.
Specifically, it is defined as 
$ \{ \PP: {\rm{W}}(\PP, \hat{\PP}) \leq r \}$,
where $r$ is the prescribed radius, and the Wasserstein distance is defined as
\begin{align*}
    {{\rm{W}}}(\mathbb{P},\hat{\BP}) = \inf_{ \gamma \in \Gamma(\mathbb{P},\hat{\BP})} \mathbb{E}_{(\xi,\zeta)\sim\gamma} [c(\xi,\zeta)].
\end{align*}
Here, $\Gamma({\PP},\hat{\PP})$ denotes the set of joint distributions whose marginal distributions are $\PP$ and $\hat{\PP}$, respectively, and $c(\xi,\zeta)$ represents the transportation cost from sample $\xi$ to sample $\zeta$.

In practice, the datasets usually contain outliers \cite{jyothi2020supervised,hughes2021increasing}, whose transportation to normal data incurs a large cost of $C$. 
In such cases, Wasserstein DRO must adopt a large radius $r$ to ensure that the true distribution $\PP^*$ lies within the ambiguity set, as ${\rm{W}}(\PP, \PP^*)$ becomes large in the presence of outliers.
However, this large radius also admits many unlikely distributions, thereby making the model overly conservative.

To deal with outliers, inspired by \cite{wangoutlier}, we consider the following ambiguity set 
\begin{align}\label{eq:ambiguity}
    \mathcal{A} = \{ \PP: {\rm{UW}}(\PP || \hat{\PP}) \leq r \},
\end{align} 
where the unbalanced Wasserstein distance is defined as
\begin{align}\label{eq:def:UW}
    {{\rm{UW}}}(\mathbb{P} || \hat{\mathbb{P}}) = \inf_{\bar{\mathbb{P}},\gamma \in \Gamma(\mathbb{P},\bar{\mathbb{P}})} \left\{ \mathbb{E}_{(\xi,\zeta)\sim \gamma } [c(\xi,\zeta)] + \beta \rm{D_{KL}}(\bar{\mathbb{P}} || \hat{\mathbb{P}})\right\}.
\end{align}
Here, $c$ is the transportation cost function as defined in the Wasserstein distance, $\beta$ is a weight scalar, and $\Gamma({\PP},\bar{\PP})$ denotes the set of joint distributions whose marginal distributions are $\PP$ and $\bar{\PP}$, respectively. 
Compared to the traditional Wasserstein distance, this formulation introduces flexibility by allowing the marginal distribution $\hat{\BP}$ to adjust to $\bar{\BP}$. This adjustment is not free and induces a penalization cost based on the ${\text{KL}}$ divergence. Since the ${\text{KL}}$ divergence is not sensitive to the support of distributions, it is the key to mitigating the influence of outliers. The penalization cost is then aggregated with the Wasserstein cost between $\BP$ and $\bar{\BP}$.
With the following lemma, we show that the ${\rm{UW}}$ distance between the true distribution $\PP^*$ and contaminated distribution $\hat{\PP}$ can be effectively bounded. The proof is given in the Appendix.
\begin{lemma}\label{lemma:UW}
For any probability $\bar{\PP} \in \mathcal{P}(\Xi)$, we have
\begin{align}
    \rm{UW}(\PP^*|| \hat{\PP}) \leq {\rm{W}}(\PP^*,\bar{\PP}) + \beta  D_{KL}(\bar{\PP}|| \hat{\PP}).
\end{align}
\end{lemma}
Suppose that $\hat{\PP}_{\mathrm{clean}}$ denotes the clean distribution obtained from $\hat{\PP}$ after removing all outliers. Based on Lemma~\ref{lemma:UW}, we have
\begin{align*}
    \rm{UW}(\PP^*|| \hat{\PP}) \leq W(\PP^*,\hat{\PP}_{clean}) + \beta  D_{KL}(\hat{\PP}_{clean}|| \hat{\PP}).
\end{align*}
Intuitively, the ${\rm{UW}}$ distance between $\PP^*$ and $\hat{\PP}$ is decomposed into two components: a Wasserstein distance between $\PP^*$ and $\rm{\hat{\PP}_{clean}}$, and a ${\rm{KL}}$ penalization cost that quantifies the mismatch between $\rm{\hat{\PP}_{clean}}$ and $\hat{\PP}$. Since the $\rm{KL}$ divergence measures only relative differences, $\rm{D_{KL}}(\hat{\PP}_{clean}|| \hat{\PP})$ remains small. As a result, a small radius $r$ in \eqref{eq:ambiguity} can effectively include the distributions of interest.

\subsection{Outlier-Robust Federated Learning}

We consider a federated learning (FL) setting in which $N$ clients collaboratively train a common model parameterized by $\theta$. Each client $i \in \{1,\dots,N\}$ possesses a local empirical data distribution $\hat{\mathbb{P}}_i$. For a given set of nonnegative client weights $\lambda = (\lambda_1,\dots,\lambda_N)$ satisfying $\sum_{i=1}^N \lambda_i = 1$, the aggregated empirical distribution is defined as $\hat{\PP}_{\lambda} = \sum_i \lambda_i \hat{\PP}_i$.

In the presence of distribution shifts and outliers, $\hat{\mathbb{P}}_{\lambda}$ often provides a poor approximation of the true underlying distribution. 
To address this, we construct an ambiguity set based on the $\rm{UW}$ distance \eqref{eq:def:UW}. 
The $\mathrm{UW}$ distance relaxes the hard marginal constraint, thereby allowing the ambiguity set to contain a broader set of distributions, including clean distributions.
However, this ambiguity set also includes contaminated distributions that are close to $\hat{\mathbb{P}}_{\lambda}$ under the $\mathrm{UW}$ distance. Such contaminated distributions can be detrimental to model performance.
To mitigate this issue, we incorporate prior knowledge about potential outliers, which is necessary to design a provably model selection as discussed in \cite{zhai2021doro,nietert2024outlier}. Specifically, we introduce an outlier scoring function $h(\xi)$ that assigns larger values to samples suspected to be outliers.

The design of 
$h(\xi)$ is application-dependent and may be informed by domain expertise or 
statistical diagnostics. For example, in medical applications, $h(\xi)$ could 
encode thresholds for biomarker values outside physiologically plausible 
ranges. Alternatively, $h(\xi)$ may capture specific feature characteristics 
reflective of the target testing distribution. For example, when training a medical classification model intended for young individuals, samples from older individuals that exhibit distinct physiological patterns may be treated as outliers. In this case, $h(\xi)$ could be designed to capture age-related characteristics, assigning larger values to samples that lie outside the targeted age range.

We then consider the following optimization problem
\begin{align*}
    \min_{\theta} \sup_{\PP: {\rm{UW}}( \PP ||\hat{\PP}_{\lambda})\leq r } \EE_{\xi \sim \PP}[L(\theta,\xi)],
\end{align*}
where $L(\theta,\xi) = l(\theta,\xi) - h(\xi)$. For any distribution $\BP$ that contains outliers, $\EE_{\xi\sim \BP}[h(\xi)]$ is large and thus $\BP$ is less likely to attain the supremum.

\begin{figure}[t] 
\begin{center}
\centerline{\includegraphics[width=0.6\columnwidth]{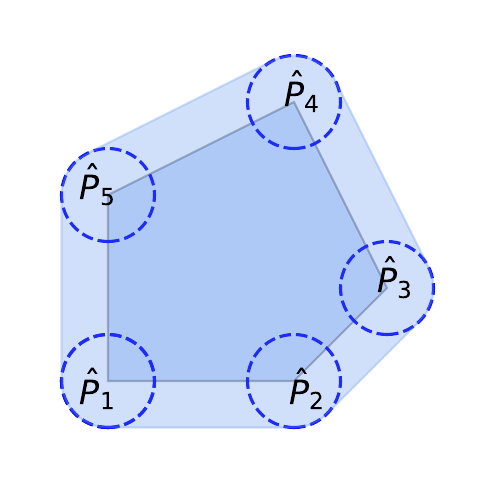}}
\caption{Illustration of the ambiguity set for the considered problem \eqref{eq:problem}.}
\label{fig_illus}
\end{center}
\end{figure}

As shown in \cite{mohri2019agnostic}, using a fixed value of $\lambda$ may result in a model that is biased toward certain clients. Therefore, we refine the above formulation to the following optimization problem:
\begin{align}\label{eq:problem}
    \min_{\theta} \max_{\lambda \in \Lambda} \sup_{\PP: {\rm{UW}}( \PP ||\hat{\PP}_{\lambda})\leq r } \EE_{\xi \sim \PP}[L(\theta,\xi)],
\end{align}
where $\Lambda = \{ \lambda: \lambda_i\geq 0, \; \sum_{i=1}^N \lambda_i = 1\}$. 
Geometrically, the overall ambiguity set in \eqref{eq:problem} can be interpreted as the Minkowski sum of the UW ambiguity sets $\{\mathbb{P} : {\rm UW}(\mathbb{P}\,\|\,\hat{\mathbb{P}}_{\lambda}) \le r\}$ over all $\lambda \in \Lambda$. See Fig.~\ref{fig_illus} for an illustration.

The formulation 
\eqref{eq:problem} captures data heterogeneity, distribution shifts, and outlier resilience.
Our goal in this paper is to solve \eqref{eq:problem} in a decentralized manner suitable for FL.

\section{Tractable Reformulation}\label{sec:3_result}
The problem \eqref{eq:problem} exhibits a $\min-\max-\max$ structure, where the innermost maximization is carried out over a constrained set of probability measures in an infinite-dimensional space, making the decentralized training intractable.
Motivated by~\cite{sinha2017certifying}, we replace the hard constraint in the DRO formulation \eqref{eq:problem} with a soft penalization, leading to the Lagrangian penalty form
\begin{align}\label{eq:Lagrangian}
    \min_{\theta} \max_{\lambda \in \Lambda} \sup_{\PP  } \EE_{\xi \sim \PP}[L(\theta,\xi)] - \rho \text{UW}( \PP ||\hat{\PP}_{\lambda}).
\end{align}
where $\rho>0$ is a penalty coefficient. We note that \eqref{eq:Lagrangian} still has a $\min-\max-\max$ structure and continues to capture data heterogeneity, distribution shifts, and outlier resilience. However, it is simplified because the innermost maximization becomes unconstrained.
In what follows, we establish strong duality for \eqref{eq:Lagrangian} and show that it enables decentralized training. The proof is provided in the Appendix. 
\begin{proposition}\label{prop:reformulation}
The problem formulation \eqref{eq:Lagrangian} is equivalent to 
\begin{align}\label{eq:FL:problem}
    \min_{\theta} \max_{\lambda \in \Lambda} \sum_i \lambda_i \EE_{\zeta \sim \hat{\PP}_i} \Big[ e^{\sup_{\xi} \left\{ L(\theta,\xi) - \rho c(\zeta,\xi) \right\} / (\rho \beta)} \Big] .
\end{align}
\end{proposition}

We observe that the expression in \eqref{eq:FL:problem} decomposes as a weighted sum of client-specific terms, making it naturally amenable to distributed training 
in FL environments.

Although the formulation  \eqref{eq:Lagrangian} regulates the degree of robustness through the penalty coefficient $\rho$, we show in the following proposition that it still provides robustness certificates. 
\begin{proposition}\label{prop:robust:guarantee}
Fix $\rho >0$ and $\lambda$. Let $$\PP^*_{\lambda} = {\rm{arg max}}_{\PP} \{ \EE_{\xi \sim \PP}[L(\theta,\xi)] - \rho {\rm{UW}}( \PP ||\hat{\PP}_{\lambda})\},$$ and define $\hat{r}_{\lambda} = \rm{UW}(\PP^*_{\lambda}|| \hat{\PP}_{\lambda})$.
Then, we have
\begin{align}
    & \sup_{\PP: {\rm{UW}}( \PP ||\hat{\PP}_{\lambda})\leq \hat{r}_{\lambda} } \EE_{\xi \sim \PP}[L(\theta,\xi)]  \nonumber \\
    &\leq \rho \hat{r}_{\lambda} + \rho \beta \log \EE_{\zeta \sim \hat{\PP}_{\lambda}}[ e^{f(\theta,\zeta)/(\rho \beta)}] \nonumber \\
    & = \EE_{\PP^*_{\lambda}} [L(\theta,\xi)],
\end{align}
where $f(\theta,\zeta) := \sup_{\xi \in \Xi} L(\theta,\xi) - \rho c(\zeta,\xi)$.
\end{proposition}
Proposition~\ref{prop:robust:guarantee} implies that for any $\rho > 0$, 
the maximizer $\mathbb{P}^*_{\lambda}$ of the Lagrangian penalty problem 
also maximizes the original DRO problem over the ambiguity set 
$\{\mathbb{P} : {\rm UW}(\mathbb{P} \,\|\, \hat{\mathbb{P}}_{\lambda}) 
\le \hat{r}_{\lambda} \}$, where $\hat{r}_{\lambda}$ is determined 
by $\mathbb{P}^*_{\lambda}$. In particular, solving~\eqref{eq:Lagrangian} 
ensures distributional robustness with respect to this induced ambiguity set.
\section{Algorithm}\label{sec:4_alg}
\subsection{Algorithm Design}

\begin{algorithm}[t]
\caption{Distributionally Outlier-Robust Federated Learning} \label{alg:FL}
\begin{algorithmic}[1]
    \REQUIRE Sampling distribution $\hat{\PP}$, constraint sets $\Theta$ and $\Lambda$, step size $\eta_{\theta}$ and $\eta_{\lambda}$
    \STATE Server broadcasts the function $h$ to each client
    \FOR{$ t=1,\ldots,T$}
        \FOR{$\text{client}$ $i=1,\ldots,N$}
            \STATE Sample $\hat{\zeta}_{i,t}$ from $\hat{\PP}_{i}$
            \STATE Find an $\epsilon$-approximate maximizer $z_{i,t}$ of $\text{argmax}_{\xi \in \Xi} L(\theta_t, \xi) - \rho c(\xi, \hat{\zeta}_{i,t})$
            \STATE Construct gradient estimate $g_{i,t}^{\theta}$ by \eqref{eq:alg:git_theta} 
            \STATE Construct gradient estimate $g_{i,t}^{\lambda}$ by \eqref{eq:alg:git_lambda}
            \STATE Update $\theta_{i,t+1}$ by \eqref{eq:alg:update_theta}
            \STATE Client $i$ sends $\theta_{i,t+1}$, $g_{i,t}^{\lambda}$ back to server
        \ENDFOR
        \STATE  Server updates $\theta_{t+1}$ by \eqref{eq:alg:server:theta} 
        \STATE Server updates $\lambda_{t+1}$ by \eqref{eq:alg:server:lmbda} 
        \STATE Server broadcasts $\theta_{t+1}$ and $\lambda_{t+1}$ to each client
    \ENDFOR
\end{algorithmic}
\end{algorithm}

To solve the problem ~\eqref{eq:FL:problem}, we propose the Distributionally Outlier-Robust Federated Learning (DOR-FL) algorithm, summarized in Algorithm~\ref{alg:FL}. 
We assume that the server specifies the outlier scoring function $h(\xi)$ and broadcasts it to all clients. Each client then participates in training the common model through iterative communication with the server.

Specifically, at each time step $t$, each client $i$ draws a sample $\hat{\zeta}_{i,t}$ from its empirical distribution $\hat{\mathbb{P}}_i$ and solves
\begin{align}\label{eq:inner_max}
    \text{argmax}_{\xi \in \Xi} \; L(\theta_t, \xi) - \rho \, c(\xi, \hat{\zeta}_{i,t}),
\end{align}
obtaining an approximate maximizer $z_{i,t}$ such that 
${\rm dist}(z_{i,t}, X_{i,t}^*) \leq \epsilon$, where $\rm{dist}$ denotes the distance from a point to the set, and $X_{i,t}^*$ denotes the set of exact maximizers. This can be achieved efficiently via gradient-based methods when $L(\theta,\xi)$ is concave in $\xi$ and $c(\xi,\zeta)$ is (strongly) convex in $\xi$.
Given $z_{i,t}$, client $i$ computes a stochastic gradient estimator with respect to $\theta$:
\begin{align}\label{eq:alg:git_theta}
    g_{i,t}^{\theta}
    = \exp\!\left( \frac{L(\theta_t,z_{i,t}) - \rho c(z_{i,t},\hat{\zeta}_{i,t})}{\rho \beta} \right)
      \frac{\nabla_{\theta} L(\theta_t,z_{i,t})}{\rho \beta},
\end{align}
and updates locally via
\begin{align}\label{eq:alg:update_theta}
    \theta_{i,t+1} = \theta_t - \eta_{\theta} \, g_{i,t}^{\theta},
\end{align}
where $\eta_{\theta}$ is the step size for updating $\theta$.
Similarly, it computes a gradient estimator with respect to $\lambda$:
\begin{align}\label{eq:alg:git_lambda}
    g_{i,t}^{\lambda}
    = \exp\!\left( \frac{L(\theta_t,z_{i,t}) - \rho c(z_{i,t},\hat{\zeta}_{i,t})}{\rho \beta} \right).
\end{align}
Each client $i$ then broadcasts $\theta_{i,t+1}$, $g_{i,t}^{\lambda}$ to the server. The server updates the global parameters:
\begin{align}\label{eq:alg:server:theta}
    \theta_{t+1} = {\rm{Proj}}_{\Theta}(\sum_{i=1}^N \lambda_{i,t} \theta_{i,t+1} ),
\end{align}
and 
\begin{align}\label{eq:alg:server:lmbda}
    \lambda_{t+1} ={\rm{Proj}}_{\Lambda} \Big( \lambda_t + \eta_{\lambda} g_t^{\lambda} \Big),
\end{align}
where $g_t^{\lambda} = (g_{1,t}^{\lambda},\ldots, g_{N,t}^{\lambda})$, and $\eta_{\lambda}$ is the step size for updating $\lambda$. The updated $(\theta_{t+1}, \lambda_{t+1})$ are then broadcast to all clients.

\subsection{Convergence Analysis}
In this section, we provide the convergence analysis for the proposed algorithm. For convenience, define 
$$H(\theta,\lambda) = \EE_{\zeta \sim \hat{\PP}_{\lambda}} \Big[ e^{ f(\theta,\zeta) / (\rho \beta)} \Big]. $$
Our convergence analysis requires the following assumptions.

\begin{assumption}\label{assump:L}
$L(\theta,\xi)$ is convex in $\theta$ for every $\xi \in \Xi$ and concave in $\xi$. 
\end{assumption}

\begin{assumption}\label{assump:diameter}
The domains are bounded such that
$\max_{\theta_1,\theta_2 \in \Theta} \left\|\theta_1 - \theta_2 \right\| \leq D_{\theta}$, $\max_{\lambda_1,\lambda_2 \in \Lambda} \left\|\lambda_1 - \lambda_2 \right\| \leq D_{\lambda}$. 
\end{assumption}

\begin{assumption}\label{assump:bound}
The loss function $L$ satisfies that
$| L(\theta,\xi)| \leq B_1$ and $\left\| \nabla_{\theta} L(\theta,\xi)\right\| \leq B_2$.
\end{assumption}

\begin{assumption}\label{assump:c}
$c(\xi,\zeta)$ is strongly convex in $\xi$.
\end{assumption}
\begin{assumption}\label{assump:DRO}
$\nabla_{\theta} L(\theta,\xi)$ is $L_{\theta \xi}$-Lipschitz continuous in $\xi$, and $L(\theta,\xi) - \rho c(\xi,\zeta)$ is differentiable and $L_{\xi}$-smooth in $\xi$ for every $\zeta \in \Xi$.
\end{assumption}

Assumption~\ref{assump:L} is widely used in both DRO and FL literature \cite{deng2020distributionally,mohri2019agnostic}. Assumptions~\ref{assump:diameter} and \ref{assump:bound} are standard in FL studies \cite{mohri2019agnostic}. Assumption~\ref{assump:c} holds in many cases, such as the quadratic transportation cost $c(\xi,\zeta) = \frac{1}{2}\| \xi - \zeta\|_2^2$. Assumption~\ref{assump:DRO} is also standard in the DRO literature \cite{sinha2017certifying, deng2020distributionally}.

We now present the convergence analysis of Algorithm~\ref{alg:FL}, summarized in the following theorem. The detailed proof is provided in the Appendix.

\begin{theorem}\label{thm:convergence}
Let Assumptions \ref{assump:L}-\ref{assump:DRO} hold.
Select $\eta_{\lambda} = \eta_{\theta} =\frac{1}{\sqrt{T}} $. Then, Algorithm~\ref{alg:FL} satisfies
\begin{align}
    \max_{\lambda \in \Lambda} H(\bar{\theta}, \lambda) - \min_{\theta \in \Theta} \max_{\lambda \in \Lambda} H(\theta, \lambda) = \mathcal{O}(T^{-1/2} + \epsilon),
\end{align}
where $\bar{\theta} = \frac{1}{T}\sum_{t=1}^T \theta_t$.
\end{theorem}
The parameter $\epsilon$ is the accuracy of solving the inner max problem \eqref{eq:inner_max}, and it contributes a constant offset to the optimization accuracy , which is independent of the total number of steps $T$.
The $\mathcal{O}(T^{-1/2})$ convergence rate in Theorem~\ref{thm:convergence} aligns with standard federated minimax optimization results \cite{mohri2019agnostic,deng2020distributionally}. The convergence rate of the algorithm is independent of the specific choice of the outlier scoring function $h(\xi)$, provided the stated assumptions are satisfied.

\section{Experiments}\label{sec:5_experiments}

In this section, we evaluate the performance of the proposed method in both synthetic and real-world datasets.
\subsection{Synthetic Data}
We begin by assessing the proposed distributionally robust FL framework using synthetically generated datasets. 
We consider a FL environment with $N=3$ clients. 
For client $i \in \{1,2,3\}$, the feature vectors $x \in \mathbb{R}^5$ are drawn independently from a Gaussian distribution $\mathcal{N}(\mu_i, I_5)$, where $I_5$ is the identity matrix and the mean vectors are given by
\begin{align*}
    \mu_1 = (0.0, 0.0, 0.0, 0.0, 0.0), \\
    \mu_2 = (1.0, 1.0, 0.0, 0.0, 0.0), \\
    \mu_3 = (2.0, 2.0, 0.5, 1.0, 2.0).
\end{align*}
The conditional distribution of the label $y\in\{-1,1\}$ is given by
${\rm{Prob}}(y|x) = [1+ \exp(-y (\theta_{*}^{\intercal} x))]^{-1},$
where $\theta_{*}\in \mathbb{R}^5$ is a fixed parameter vector.

To emulate heterogeneous and contaminated client data, we introduce two types of perturbations: contamination perturbations, where a $\varepsilon_i$ fraction of each client’s data points are multiplied by contamination factors to generate extreme values modeling corrupted observations; and Wasserstein shifts, where the remaining samples are perturbed by adding a small translation vector to simulate covariate shifts.
The clean test set is obtained by pooling uncontaminated samples from all clients, representing the nominal global distribution.

We consider prior knowledge on the mean of the clean distribution, denoted as $\bar{\xi}$, and design the function $h(\xi) = \rho_2 \left\| \xi - \bar{\xi}\right\|^2$. 
The value of $\bar{\xi}$ is determined by a cheap preprocessing step, same as in \cite{nietert2024outlier,wangoutlier}.

\begin{figure}[t] 
\begin{center}
\centerline{\includegraphics[width=1.0\columnwidth]{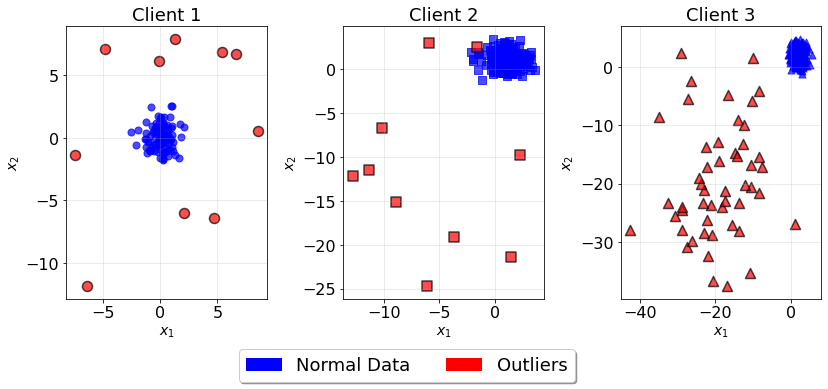}}
\caption{Visualization of the synthetic dataset.}
\label{fig:data}
\end{center}
\end{figure}

The experimental setup is as follows.
The local sample sizes for the three clients are fixed at $100$, $200$, and $500$ observations, respectively. 
The outlier contamination levels are set to $\varepsilon = (0.1, 0.05, 0.1)$, meaning that $10\%$, $5\%$, and $10\%$ of the data from Clients~1, 2, and 3, respectively, are replaced with contaminated samples.
The contamination factors, which control the magnitude of the artificially injected extreme values, are $(7, 8, 9)$ for the three clients. 
The Wasserstein shift perturbations are implemented by adding translation vectors of magnitude $(1.0, -0.5, 0.6)$ in the first feature coordinate of the respective clients' datasets. The dataset is visualized in Fig. \ref{fig:data}.

For model training, we adopt logistic regression with a maximum of $1000$ gradient descent iterations. 
The experimental results in Table~\ref{tab:syn} demonstrate that DOR-FL consistently outperforms Agnostic FL (AFL) in \cite{mohri2019agnostic}, Wasserstein FL (WAFL) in \cite{nguyen2022generalization}, and group distributionally robust FL (GDRFL) \cite{konti2025distributionally} on the overall dataset as well as on each client’s local dataset. While AFL, WAFL and GDRFL suffer from excessive conservatism in the presence of outliers, DOR-FL maintains high accuracy. These results show that our DOR-FL effectively address both geometric shifts and contamination perturbations.

\begin{figure}[t] 
\begin{center}
\centerline{\includegraphics[width=0.9\columnwidth]{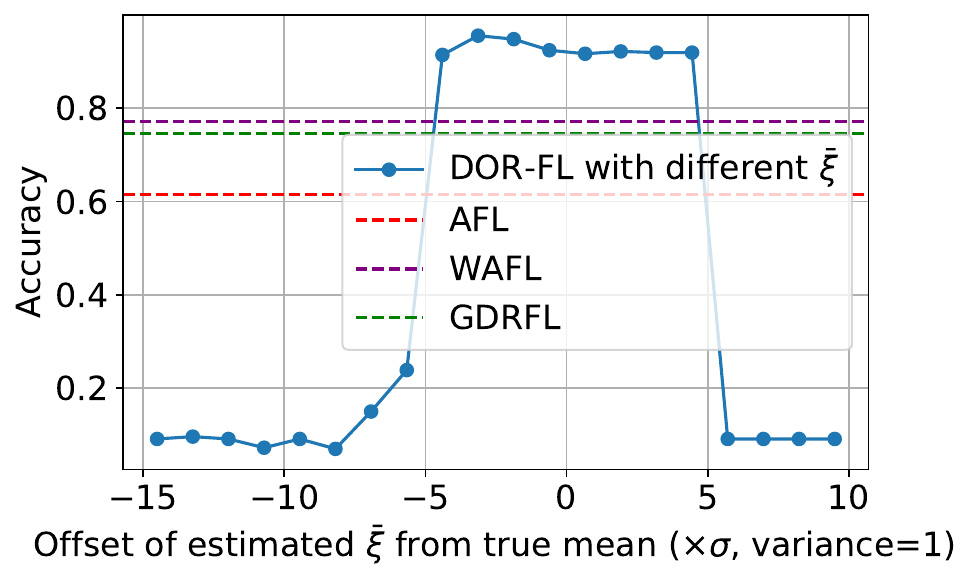}}
\caption{Classification accuracy of DOR-FL under different offsets between the estimated $\bar{\xi}$ and the true clean-data mean, expressed in multiples of the standard deviation (variance = 1).}
\label{fig:syn:diff-h}
\end{center}
\end{figure}

To further evaluate the sensitivity of DOR-FL to the choice of the prior mean $\bar{\xi}$, we vary the estimated value of $\bar{\xi}$ over a wide range and measure the resulting classification accuracy. Fig.~\ref{fig:syn:diff-h} shows that the accuracy of DOR-FL remains consistently high across offsets of up to $\pm$ 5 standard deviations from the true clean-data mean. This insensitivity to moderate or even large deviations indicates that DOR-FL does not rely on a highly accurate specification of $h(\xi)$ to maintain strong performance.

\begin{table}[t]
    \caption{Classification accuracy on synthetic datasets.}
    \centering
    \begin{tabular}{ccccc}
        \hline
        {Method } & Overall & Group 1 & Group 2 & Group 3 \\
        \hline
        AFL \cite{mohri2019agnostic}  & 61.5 & 61.9 & 64.1 & 60.4 \\
        WAFL \cite{nguyen2022generalization} & 77.1 & 69.2 & 74.0 & 79.9 \\
        GDRFL \cite{konti2025distributionally}&  74.5 & 67.0 & 73.8 & 76.3 \\ 
        DOR-FL & \textbf{95.4} & \textbf{84.6} & \textbf{91.3} & \textbf{99.2} \\
        \hline
    \end{tabular}\label{tab:syn}
\end{table}

\subsection{Real-World Data}

\begin{table}[!htbp]
    \caption{Classification accuracy and excess risk for adult income prediction on the overall UCI dataset}
    \label{tab:uci}
    \centering
    \begin{tabular}{cccc}
        \toprule
        {Method } & Accuracy\% $\uparrow$ & Excess risk $\downarrow$ \\
        \midrule
        ERM  & 83.1 & 0.37  \\
        AFL  & 81.7 & 0.44  \\
        WAFL  & 76.7 & 0.64  \\
        GDRFL & 79.5 & 0.61 \\
        DOR-FL & \textbf{84.6} & \textbf{0.34}  \\
        \bottomrule
    \end{tabular}
\end{table}


\begin{figure}[t] 
\vspace{-0.2cm}
\begin{center}
\centerline{\includegraphics[width=0.7\columnwidth]{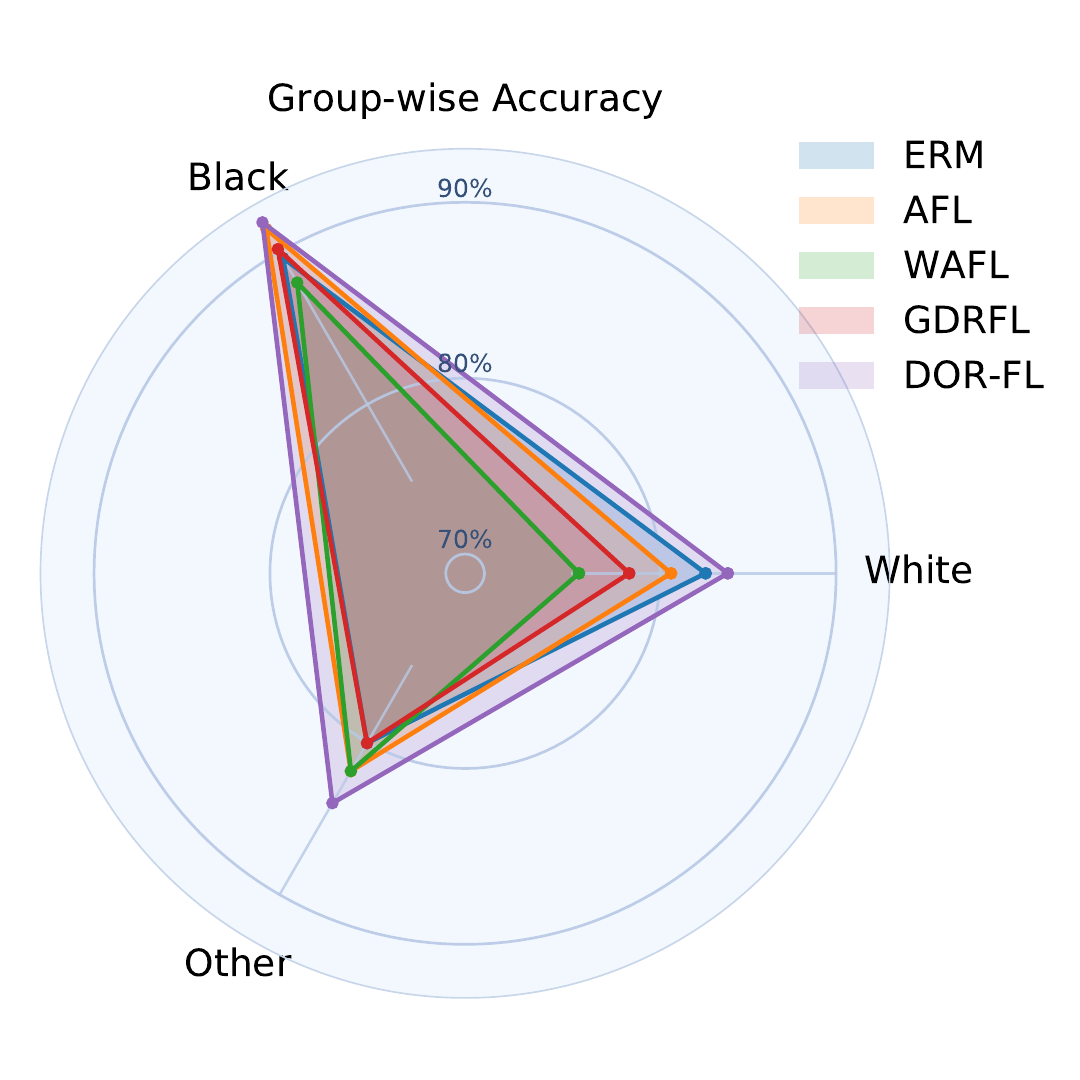}}
\caption{Radar plot of group-wise classification accuracy. DOR-FL exhibits the outermost triangle, indicating the highest accuracy across all groups. }
\label{fig:UCI_acc_group}
\end{center}
\end{figure}


\begin{figure}[t] 
\vspace{-0.5cm}
\begin{center}
\centerline{\includegraphics[width=0.7\columnwidth]{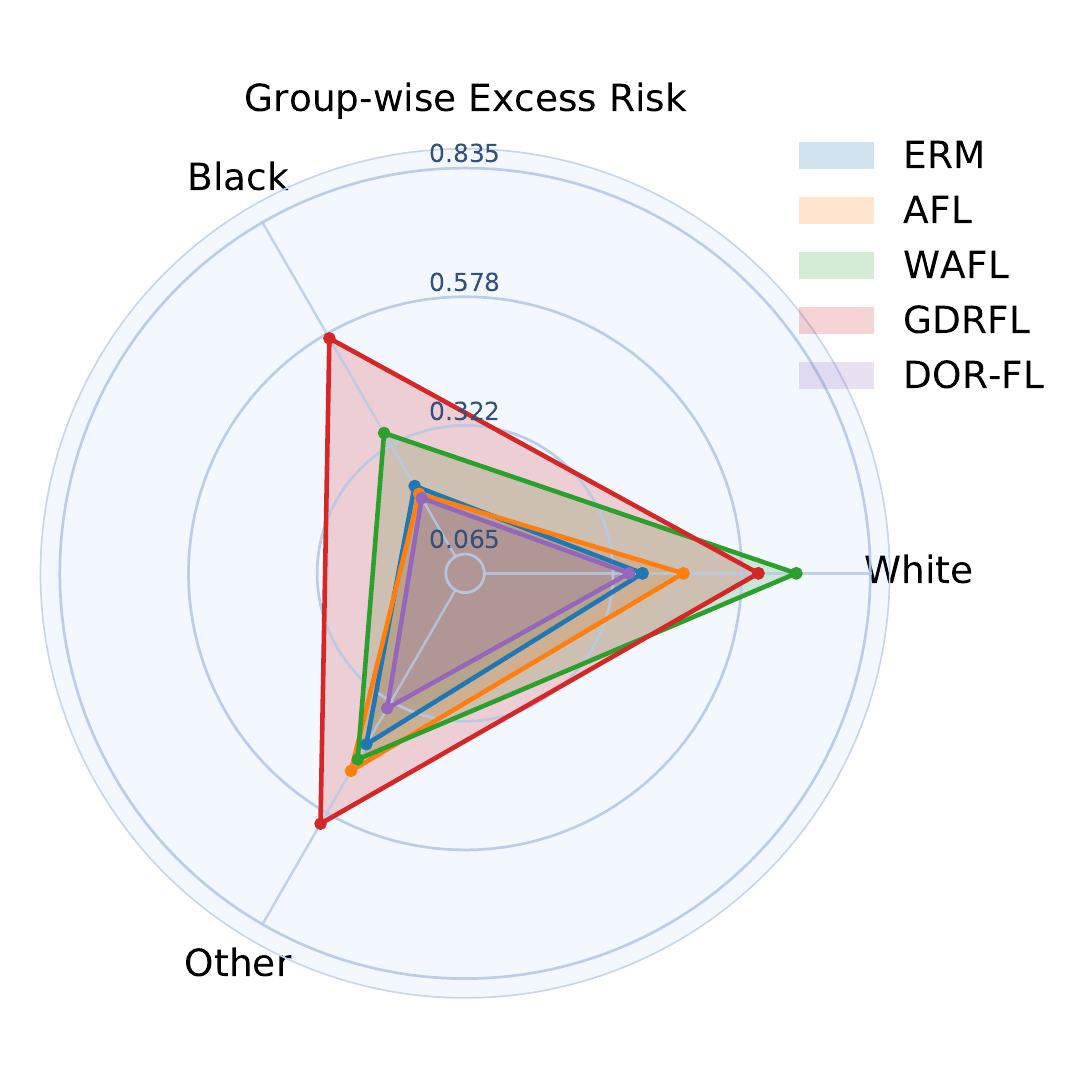}}
\caption{Radar plot of group-wise excess risk. DOR-FL exhibits the innermost triangle, indicating the lowest excess risk across all groups. }
\label{fig:UCI_risk_group}
\end{center}
\end{figure}

We further evaluate the proposed methods on the UCI Adult Income dataset \cite{adult_2}, which contains demographic and financial attributes of individuals along with a binary income label indicating whether the annual income exceeds \$50K. The dataset consists of 32,561 samples in the training split and 16,281 samples in the testing split. 
Each record includes $14$ attributes, of which $6$ are numerical (age, final weight, education years, capital gain, capital loss, and hours worked per week) and the remainder are categorical. 
To simulate a FL environment, we partition the dataset into three clients based on race: White, Black, and Other. The clients have the sample sizes of 29249, 3269, 1670, respectively.

We focus on the capital gain feature as a potential source of outliers. Empirically, most records with extremely high \texttt{capital\_gain} values 
(above \$20K) are strongly associated with high-income individuals (\texttt{income} $> \$ 50$K). 
However, we observe a small but notable set of samples where \texttt{capital\_gain} exceeds this threshold, yet the reported income label is $y=-1$, i.e., $\leq$ \$50K. These cases deviate markedly from typical patterns and are therefore treated as potential outliers in our robust optimization framework. In particular, we design the function  $h(\xi)$ that assigns larger values to samples whose features are highly indicative of high income but are labeled as low income. Let $g(\xi)$ be the standardized \texttt{capital\_gain} feature, $D$ the standardized threshold for \$20K, and $\mathbb{I}$ the indicator function. The ideal penalty is $\rho_2 \mathbb{I}\{y=-1 \ \wedge \ g(\xi) > D \}$, where $\rho_2 > 0$ controls the penalty scale. To enable smooth optimization, we replace the indicator with a sigmoid approximation, which leads to
$$h(\xi) = \rho_2 \mathbb{I}\{y = -1\}  \sigma\!\left( \frac{g(\xi) - D}{s} \right),$$
where $s > 0$ is a softness parameter. By incorporating $h$ into the distributionally robust objective, we explicitly reduce the influence of these suspected outlier points during model training.

Performance is measured using test accuracy and excess risk.
We compare our method DOR-FL with three standard FL algorithms: ERM \cite{vapnik1998statistical}, AFL \cite{mohri2019agnostic}, and WAFL \cite{nguyen2022generalization}. 
All methods are implemented with the same logistic regression model and tuned optimally.

The overall accuracy and excess risk of all methods are presented in Table~\ref{tab:uci}. 
We observe that DOR-FL achieves the highest overall accuracy and the lowest excess risk, outperforming all the baseline methods. 
The group-wise accuracy results are shown in Fig.~\ref{fig:UCI_acc_group}, where DOR-FL consistently achieves the best performance across all racial groups. 
While ERM shows reasonable performance for the majority group (White), its performance on the minority group (Other) is worse. 
DOR-FL narrows this gap, improving minority group accuracy without sacrificing overall performance.
The group-wise excess risk is presented in Fig.~\ref{fig:UCI_risk_group}, which further confirms that DOR-FL not only reduces disparity between groups but also improves the overall performance.
In summary, these results demonstrate that incorporating outlier mitigation into distributionally robust FL substantially improves both global performance and fairness across demographic groups.

\section{Conclusion}\label{sec:6_conclusion}
In this work, we proposed a distributionally robust FL framework that explicitly accounts for distribution shifts and outlier contamination through an unbalanced Wasserstein distance. Theoretically, we established both convergence guarantees and robustness certificates for DOR-FL. Practical experiments on synthetic and real-world datasets demonstrated that DOR-FL consistently outperforms existing robust FL baselines, achieving higher overall accuracy, lower excess risk, and improved fairness across demographic groups.

Future research directions include developing communication-efficient variants of DOR-FL to further reduce transmission overhead in large-scale networks and exploring scalable algorithms for high-dimensional applications.

\section*{Appendix}

\subsection{Proof of Lemma~\ref{lemma:UW}}
Following the definition in \eqref{eq:def:UW}, we have 
\begin{align*}
    &{{\rm{UW}}}(\PP^* || \hat{\mathbb{P}})  \nonumber \\
    &= \inf_{\bar{\mathbb{P}},\gamma \in \Gamma(\PP^*,\bar{\mathbb{P}})} \left\{ \mathbb{E}_{(\xi,\zeta)\sim \gamma } [c(\xi,\zeta)] + \beta \rm{D_{KL}}(\bar{\mathbb{P}} || \hat{\mathbb{P}})\right\} \nonumber \\
    & \leq \inf_{\gamma \in \Gamma(\PP^*,\bar{\mathbb{P}})} \left\{ \mathbb{E}_{(\xi,\zeta)\sim \gamma } [c(\xi,\zeta)] + \beta \rm{D_{KL}}(\bar{\mathbb{P}} || \hat{\mathbb{P}})\right\} \nonumber \\
    & = {\rm{W}}(\PP^*,\bar{\PP}) + \beta  \rm{D_{KL}}(\bar{\PP}|| \hat{\PP}).
\end{align*}
The proof is complete.

\subsection{Proof of Proposition~\ref{prop:reformulation}}

The proof follows directly from Theorem 2 in \cite{wangoutlier}. We have 
\begin{align*}
    &  \max_{\lambda} \sup_{\PP} \EE_{\xi\sim \PP }[L(\theta,\xi)] - \rho \text{UW}( \PP ||\hat{\PP}_{\lambda}) \nonumber \\ 
    \Leftrightarrow  &  \max_{\lambda} \rho \beta \log \EE_{\zeta \sim \hat{\PP}_{\lambda}} \Big[ e^{\sup_{\xi} \left\{ L(\theta,\xi) - \rho c(\zeta,\xi) \right\} / (\rho \beta)} \Big]     \nonumber \\
    \Leftrightarrow  &  \max_{\lambda} \EE_{\zeta \sim \hat{\PP}_{\lambda}} \Big[ e^{\sup_{\xi} \left\{ L(\theta,\xi) - \rho c(\zeta,\xi) \right\} / (\rho \beta)} \Big]     \nonumber \\
    \Leftrightarrow  & \max_{\lambda} \sum_i \lambda_i \EE_{\zeta \sim \hat{\PP}_i} \Big[ e^{\sup_{\xi} \left\{ L(\theta,\xi) - \rho c(\zeta,\xi) \right\} / (\rho \beta)} \Big].
\end{align*}
The proof ends.

\subsection{Proof of Proposition~\ref{prop:robust:guarantee}}
By virtue of Lemma 1 in \cite{wangoutlier}, the following weak duality holds:
\begin{align}
    & \sup_{\PP: {\rm{UW}}( \PP ||\hat{\PP}_{\lambda})\leq \hat{r}_{\lambda} } \EE_{\xi \sim \PP}[L(\theta,\xi)]  \nonumber \\
    &\leq \inf_{\alpha \geq 0} \big\{ \alpha \hat{r}_{\lambda} + \alpha \beta \log \EE_{\zeta \sim \hat{\PP}_{\lambda}}[ e^{f(\theta,\zeta)/(\alpha \beta)}] \big\}\nonumber \\
    &\leq  \rho \hat{r}_{\lambda} + \rho \beta \log \EE_{\zeta \sim \hat{\PP}_{\lambda}}[ e^{f(\theta,\zeta)/(\rho \beta)}],
\end{align}
where the last inequality is obtained by setting $\alpha = \rho$. The first inequality is proved. For the equality, we use the strong duality from Theorem 2 in \cite{wangoutlier}:
\begin{align}\label{eq:pf:prop2-1}
    \sup_{\PP} \EE_{\xi\sim \PP }[L(\theta,\xi)] - \rho \text{UW}( \PP ||\hat{\PP}_{\lambda}) \nonumber \\ 
    = \rho \beta \log \EE_{\zeta \sim \hat{\PP}_{\lambda}} \Big[ e^{ f(\theta,\zeta) / (\rho \beta)} \Big].
\end{align}
Since $\PP^*_{\lambda} = {\rm{arg max}}_{\PP} \{ \EE_{\xi \sim \PP}[L(\theta,\xi)] - \rho {\rm{UW}}( \PP ||\hat{\PP}_{\lambda})\}$, we have
\begin{align}\label{eq:pf:prop2-2}
    &\sup_{\PP  } \EE_{\xi \sim \PP}[L(\theta,\xi)] - \rho \text{UW}( \PP ||\hat{\PP}_{\lambda}) \nonumber \\
    & = \EE_{\xi \sim \PP^*_{\lambda}}[L(\theta,\xi)] - \rho \text{UW}( \PP^*_{\lambda} ||\hat{\PP}_{\lambda}) \nonumber \\
    & = \EE_{\xi \sim \PP^*_{\lambda}}[L(\theta,\xi)] - \rho \hat{r}_{\lambda}.
\end{align}

Combining \eqref{eq:pf:prop2-1} and \eqref{eq:pf:prop2-2}, we have
\begin{align*}
    \EE_{\xi \sim \PP^*_{\lambda}}[L(\theta,\xi)] = \rho \hat{r}_{\lambda} + \rho \beta \log \EE_{\zeta \sim \hat{\PP}_{\lambda}} \Big[ e^{ f(\theta,\zeta) / (\rho \beta)} \Big].
\end{align*}
The proof is complete. 

\subsection{Proof of Theorem 1}
Since $H(\theta,\lambda)$ is convex in $\theta$ and concave in $\lambda$, and $\Lambda$, $\Theta$ are convex compact sets, by the general minmax theorem \cite{sion1958general}, we have
\begin{align}\label{eq:centra:t1}
    &\max_{\lambda \in \Lambda} H(\bar{\theta}, \lambda) - \min_{\theta \in \Theta} \max_{\lambda \in \Lambda} H(\theta, \lambda) \nonumber \\
    &= \max_{\lambda \in \Lambda} H(\bar{\theta}, \lambda) - \max_{\lambda \in \Lambda}\min_{\theta \in \Theta}  H(\theta, \lambda) \nonumber \\
    & \leq \max_{\lambda \in \Lambda} H(\bar{\theta}, \lambda) - \min_{\theta \in \Theta}  H(\theta, \bar{\lambda}) \nonumber \\
    & = \max_{\lambda \in \Lambda, \theta\in \Theta} \Big\{ H(\bar{\theta}, \lambda) - H(\theta, \bar{\lambda})  \Big\} \nonumber \\
    & = \max_{\lambda \in \Lambda, \theta\in \Theta} \Big\{ H(\bar{\theta}, \lambda) - \frac{1}{T} \sum_{t=1}^T H(\theta, \lambda_t)  \Big\} \nonumber \\
    & \leq  \max_{\lambda \in \Lambda, \theta\in \Theta} \Big\{ \frac{1}{T} \sum_{t=1}^T H(\theta_t, \lambda) - \frac{1}{T} \sum_{t=1}^T H(\theta, \lambda_t)  \Big\} \nonumber \\
    & = \frac{1}{T} \max_{\lambda \in \Lambda, \theta\in \Theta} \Big\{ \sum_{t=1}^T \Big( H(\theta_t, \lambda) - H(\theta, \lambda_t) \Big) \Big\},
\end{align}
where third equality follows since $H(\theta,\lambda)$ is linear in $\lambda$, and the last inequality follows from the convexity of $H(\theta,\lambda)$ in $\theta$.
It follows that
\begin{align}\label{eq:centra:t2}
    &H(\theta_t, \lambda) - H(\theta, \lambda_t) \nonumber \\
    &= H(\theta_t, \lambda) - H(\theta_t, \lambda_t) + H(\theta_t, \lambda_t) - H(\theta, \lambda_t) \nonumber \\
    & \leq \left\langle \lambda - \lambda_t ,\nabla_{\lambda} H(\theta_t,\lambda_t) \right\rangle + \left\langle \theta_t - \theta , \nabla_{\theta} H(\theta_t,\lambda_t) \right\rangle \nonumber \\
    & = \left\langle \lambda - \lambda_t , g_t^{\lambda} \right\rangle + \left\langle \lambda - \lambda_t ,  \nabla_{\lambda} H(\theta_t,\lambda_t) -g_t^{\lambda}\right\rangle \nonumber \\
    &\quad + \left\langle \theta_t - \theta , g_t^{\theta} \right\rangle + \left\langle \theta_t - \theta , \nabla_{\theta} H(\theta_t,\lambda_t) - g_t^{\theta} \right\rangle .
\end{align}

Substituting \eqref{eq:centra:t2} into \eqref{eq:centra:t1}, we have
\begin{align}\label{eq:centra:t3}
    &\max_{\lambda \in \Lambda} H(\bar{\theta}, \lambda) - \min_{\theta \in \Theta} \max_{\lambda \in \Lambda} H(\theta, \lambda) \nonumber \\
    &\leq \frac{1}{T}\Big( \max_{\theta \in \Theta} \sum_{t=1}^T \left\langle \theta_t - \theta , g_t^{\theta} \right\rangle + \max_{\lambda \in \Lambda} \sum_{t=1}^T \left\langle \lambda - \lambda_t , g_t^{\lambda} \right\rangle \nonumber \\
    & +  \max_{\theta \in \Theta} \sum_{t=1}^T\left\langle \theta_t - \theta , \nabla_{\theta} H(\theta_t,\lambda_t) - g_t^{\theta} \right\rangle \nonumber \\
    & + \max_{\lambda \in \Lambda} \sum_{t=1}^T \left\langle \lambda - \lambda_t ,  \nabla_{\lambda} H(\theta_t,\lambda_t) -g_t^{\lambda}\right\rangle \Big).
\end{align}
We deal with the terms one by one. By virtue of the update rule \eqref{eq:alg:server:theta},  for any $\theta \in \Theta$, we have
\begin{align}\label{eq:centra:t4}
    &\left\| \theta_{t+1} - \theta \right\|^2 \nonumber \\
    &\leq \left\| \theta_t - \eta_{\theta} \sum_{i=1}^N \lambda_{i,t} g_{i,t}^{\theta} - \theta \right\|^2 \nonumber \\
    & = \left\| \theta_{t} - \theta \right\|^2 + \eta_{\theta}^2 \left\| \sum_{i=1}^N \lambda_{i,t} g_{i,t}^{\theta} \right\|^2  - 2 \eta_{\theta} \left\langle\theta_{t} - \theta, \sum_{i=1}^N \lambda_{i,t} g_{i,t}^{\theta} \right\rangle,
\end{align}
where the first inequality follows since the projection operator is non-expansive. It follows that
\begin{align}\label{eq:centra:t5}
    &\sum_{t=1}^T \left\langle \theta_t - \theta , g_t^{\theta} \right\rangle \nonumber \\
    & \leq \frac{1}{2 \eta_{\theta}} \sum_{t=1}^T \Big(\left\| \theta_{t} - \theta \right\|^2 - \left\| \theta_{t+1} - \theta \right\|^2 + \eta_{\theta}^2 \left\| \sum_{i=1}^N \lambda_{i,t} g_{i,t}^{\theta} \right\|^2\Big) \nonumber \\
    &\leq \frac{ \left\| \theta_{1} - \theta \right\|^2}{2\eta_{\theta}} + \frac{\eta_{\theta}}{2} \sum_{t=1}^T \left\| \sum_{i=1}^N \lambda_{i,t} g_{i,t}^{\theta} \right\|^2 \nonumber \\
    &\leq \frac{D_{\theta}^2}{2\eta_{\theta}} + \frac{\eta_{\theta}}{2} \sum_{t=1}^T  \Big( \sum_{i=1}^N \lambda_{i,t} \left\|g_{i,t}^{\theta} \right\| \Big)^2 \nonumber \\
    &\leq \frac{D_{\theta}^2}{2\eta_{\theta}} + \frac{\eta_{\theta} T C_0^2}{2},
\end{align}
where the third inequality follows from Assumption~\ref{assump:diameter}. The last inequality follows since $\left\|g_{i,t}^{\theta} \right\| = \left\| \exp\left({ (L(\bar{\theta}_t,z_{i,t})- \rho c(z_{i,t},\hat{\xi}_{i,t}) ) /(\rho \beta)} \right) \frac{1}{\rho \beta} \nabla_{\theta} L(\bar{\theta}_t,z_{i,t})\right\| \leq  \frac{B_2}{\rho \beta} e^{\frac{B_1}{\rho \beta}}   := C_0$.

Since $L(\theta,\xi) - \rho c(\xi,\zeta)$ is differentiable and strongly concave in $\xi$, it follows from Danskin’s theorem \cite{bertsekas1997nonlinear} that $f(\theta,\zeta)$ is convex and differentiable in $\theta$. Moreover, the gradient of $f(\theta,\zeta)$ with respect to $\theta$ is given by $\nabla_{\theta} f(\theta,\zeta) = \nabla_{\theta} L(\theta,\xi^*(\zeta))$, where $\xi^*(\zeta) = {\rm{argmax}}_{\xi \in \Xi} L(\theta,\xi) - \rho c(\xi,\zeta)$.
Therefore, we have
\begin{align*}
    \nabla_{\theta} H(\theta_t,\lambda_t)& =  \EE_{\zeta \sim \hat{\PP}_{\lambda_t}} \Big[ e^{ f(\theta_t,\zeta) / (\rho \beta)} \frac{1}{\rho \beta} \nabla_{\theta} f(\theta_t,\zeta) \Big]
    \nonumber \\
    & = \sum_{i=1}^N \lambda_{i,t} \EE_{\zeta \sim \hat{\PP}_{i}} \Big[ e^{ f(\theta_t,\zeta) / (\rho \beta)} \frac{1}{\rho \beta} \nabla_{\theta} f(\theta_t,\zeta) \Big].
\end{align*}
Define $G_{i,t}^{\theta}(\zeta) = e^{ f(\theta_t,\zeta) / (\rho \beta)} \frac{1}{\rho \beta} \nabla_{\theta} f(\theta_t,\zeta)$. 
$G_{i,t}^{\theta}(\hat{\zeta}_{i,t})$ is an unbiased estimate of the gradient for client $i$.
The gradient $g_{i,t}^{\theta}$ of each client is biased and we define the bias error as 
\begin{align*}
    e_{i,t}^{\theta} = g_{i,t}^{\theta} - G_{i,t}^{\theta}(\hat{\zeta}_{i,t}).
\end{align*}
Denote by $z_{i,t}^*$ the unique maximizer of $L(\theta_t,\xi) - \rho c(\xi,\hat{\zeta}_{i,t})$, so we have $f(\theta_t,\hat{\zeta}_{i,t}) = L(\theta_t,z_{i,t}^*) - \rho c(z_{i,t}^*,\hat{\zeta}_{i,t})$ and $\nabla_{\theta} f(\theta_t, \hat{\zeta}_{i,t}) = \nabla_{\theta} L(\theta_t,z_{i,t}^*)$.
It follows that
\begin{align}\label{eq:centra:t7}
    & \left\| e_{i,t}^{\theta} \right\| \nonumber \\
    &= \Big\| \frac{1}{\rho \beta}  \exp \Big( \frac{ L(\theta_t, z_{i,t}) - \rho c(z_{i,t}, \hat{\zeta}_{i,t})} {\rho\beta} \Big)  \nabla_{\theta} L(\theta_t,z_{i,t})  \nonumber \\
    &\quad - \frac{1}{\rho \beta}  \exp \Big( \frac{ L(\theta_t, z_{i,t}^*) - \rho c(z_{i,t}^*, \hat{\zeta}_{i,t})} {\rho\beta} \Big)  \nabla_{\theta} L(\theta_t,z_{i,t}^*) \Big\| \nonumber \\
    & \leq \frac{1}{\rho \beta} \Big\| \exp \Big( \frac{ L(\theta_t, z_{i,t}) - \rho c(z_{i,t}, \hat{\zeta}_{i,t})} {\rho\beta} \Big) \nonumber \\
    &\quad \times\big(\nabla_{\theta} L(\theta_t,z_{i,t}) - \nabla_{\theta} L(\theta_t,z_{i,t}^*) \big) \Big\| \nonumber \\
    & \quad + \frac{1}{\rho \beta} \Big\| \nabla_{\theta} L(\theta_t,z_{i,t}^*) \Big(\exp \Big( \frac{ L(\theta_t, z_{i,t}) - \rho c(z_{i,t}, \hat{\zeta}_{i,t})} {\rho\beta} \Big) \nonumber \\
    &\quad - \exp \Big( \frac{ L(\theta_t, z_{i,t}^*) - \rho c(z_{i,t}^*, \hat{\zeta}_{i,t})} {\rho\beta} \Big) \Big) \Big\|\nonumber \\
    &\leq \frac{1}{\rho \beta} \exp(\frac{B_1}{\rho \beta}) L_{\theta \xi}\left\| z_{i,t} - z_{i,t}^*\right\| \nonumber \\
    & \quad + \frac{B_2}{\rho \beta} \Big\|  \exp \Big( \frac{ L(\theta_t, z_{i,t}) - \rho c(z_{i,t}, \hat{\zeta}_{i,t})} {\rho\beta} \Big) \nonumber \\
    &\quad - \exp \Big( \frac{ L(\theta_t, z_{i,t}^*) - \rho c(z_{i,t}^*, \hat{\zeta}_{i,t})} {\rho\beta} \Big) \Big\|\nonumber \\
    &\leq \frac{1}{\rho \beta} \exp(\frac{B_1}{\rho \beta}) L_{\theta \xi}\left\| z_{i,t} - z_{i,t}^*\right\|  +  \frac{B_2}{\rho \beta} \exp(\frac{B_1}{\rho \beta})  \times  \nonumber \\
    &\quad   \Big\| \frac{L(\theta_t, z_{i,t}^*) - \rho c(z_{i,t}^*, \hat{\zeta}_{i,t})}{\rho \beta} -  \frac{L(\theta_t, z_{i,t}) - \rho c(z_{i,t}, \hat{\zeta}_{i,t})}{\rho \beta} \Big\| \nonumber \\
    &\leq \frac{L_{\theta \xi}}{\rho \beta} \exp(\frac{B_1}{\rho \beta}) \left\| z_{i,t} - z_{i,t}^*\right\| + \frac{B_2}{\rho^2 \beta^2} \exp(\frac{B_1}{\rho \beta}) L_{\xi} \left\| z_{i,t} - z_{i,t}^*\right\|^2 \nonumber \\
    &\leq \frac{L_{\theta \xi}}{\rho \beta} \exp(\frac{B_1}{\rho \beta}) \epsilon + \frac{B_2}{\rho^2 \beta^2} \exp(\frac{B_1}{\rho \beta}) L_{\xi} \epsilon^2 \nonumber \\
    & = C_1 \epsilon + C_2 \epsilon^2,
\end{align}
where we define $C_1 = \frac{L_{\theta \xi}}{\rho \beta} \exp(\frac{B_1}{\rho \beta})$ and $C_2 = \frac{B_2}{\rho^2 \beta^2} \exp(\frac{B_1}{\rho \beta}) L_{\xi}$. The first inequality follows from the subadditivity of $\max$, the second inequality follows from Assumptions~\ref{assump:bound} and \ref{assump:DRO}, the third inequality follows since $|e^x - e^y| \leq e^B|x-y|$ when $x,y<B$. The fourth inequality follows from Assumption~\ref{assump:DRO}. The last inequality follows since $z_{i,t}$ is $\epsilon$-approximate maximizer.

Using , we have
\begin{align}\label{eq:centra:t8}
    & \max_{\theta \in \Theta} \sum_{t=1}^T\left\langle \theta_t - \theta , \nabla_{\theta} H(\theta_t,\lambda_t) - g_t^{\theta} \right\rangle \nonumber \\
    & =\max_{\theta \in \Theta}  \sum_{t=1}^T\left\langle \theta_t - \theta , \nabla_{\theta} H(\theta_t,\lambda_t) - \sum_{i=1}^N  \lambda_{i,t} G_{i,t}^{\theta}(\hat{\zeta}_{i,t}) \right\rangle \nonumber \\
    &\quad + \sum_{t=1}^T \left\langle \theta_t - \theta , \sum_{i=1}^N  \lambda_{i,t} G_{i,t}^{\theta}(\hat{\zeta}_{i,t}) - \sum_{i=1}^N  \lambda_{i,t} g_{i,t}^{\theta} \right\rangle \nonumber \\
    &\leq \max_{\theta \in \Theta}  \sum_{t=1}^T\left\langle \theta_t - \theta , \nabla_{\theta} H(\theta_t,\lambda_t) - \sum_{i=1}^N  \lambda_{i,t} G_{i,t}^{\theta}(\hat{\zeta}_{i,t}) \right\rangle \nonumber \\
    &\quad + \sum_{t=1}^T D_{\theta} \left\| \sum_{i=1}^N \lambda_{i,t} e_{i,t}^{\theta}\right\| \nonumber \\
    &\leq \max_{\theta \in \Theta}  \sum_{t=1}^T\left\langle - \theta , \nabla_{\theta} H(\theta_t,\lambda_t) - \sum_{i=1}^N  \lambda_{i,t} G_{i,t}^{\theta}(\hat{\zeta}_{i,t}) \right\rangle \nonumber \\
    &\quad + \left\langle \theta_t , \nabla_{\theta} H(\theta_t,\lambda_t) - \sum_{i=1}^N  \lambda_{i,t} G_{i,t}^{\theta}(\hat{\zeta}_{i,t}) \right\rangle  \nonumber \\
    &\quad +  T D_{\theta} (C_1 \epsilon + C_2 \epsilon^2),
\end{align}
where the last inequality follows since $\sum_{i=1}^N \lambda_{i,t}=1$ and $\left\| \sum_{i=1}^N \lambda_{i,t} e_{i,t}^{\theta}\right\| \leq \sum_{i=1}^N   \lambda_{i,t} \left\|e_{i,t}^{\theta}\right\| \leq C_1 \epsilon + C_2 \epsilon^2$.

Note that $\EE[ \sum_{i=1}^N \lambda_{i,t} G_{i,t}^{\theta}(\hat{\zeta}_{i,t})] = \nabla_{\theta} H(\theta_t,\lambda_t)$.
Besides,
Taking the expectation with respect to $\hat{\zeta}_{i,t} $ on both sides of \eqref{eq:centra:t8}, we have
\begin{align}\label{eq:centra:t9}
    & \EE \left[ \max_{\theta \in \Theta} \sum_{t=1}^T\left\langle \theta_t - \theta , \nabla_{\theta} H(\theta_t,\lambda_t) - g_t^{\theta} \right\rangle \right] \nonumber \\
    &\leq  \EE \left[ \max_{\theta \in \Theta}  \sum_{t=1}^T\left\langle - \theta , \nabla_{\theta} H(\theta_t,\lambda_t) - \sum_{i=1}^N  \lambda_{i,t} G_{i,t}^{\theta}(\hat{\zeta}_{i,t}) \right\rangle \right] \nonumber \\
    &\quad +  T D_{\theta} (C_1 \epsilon + C_2 \epsilon^2) \nonumber \\
    &\leq D_{\theta}   \EE \left\| \sum_{t=1}^T\big( \nabla_{\theta} H(\theta_t,\lambda_t) - \sum_{i=1}^N  \lambda_{i,t} G_{i,t}^{\theta}(\hat{\zeta}_{i,t}) \big)\right\| \nonumber \\
    &\quad +  T D_{\theta} (C_1 \epsilon + C_2 \epsilon^2) \nonumber \\
    &\leq D_{\theta}  \sqrt{ \EE \left\| \sum_{t=1}^T\big( \nabla_{\theta} H(\theta_t,\lambda_t) - \sum_{i=1}^N  \lambda_{i,t} G_{i,t}^{\theta}(\hat{\zeta}_{i,t}) \big)\right\|^2 } \nonumber \\
    &\quad +  T D_{\theta} (C_1 \epsilon + C_2 \epsilon^2) \nonumber \\
    &= D_{\theta}  \sqrt{ \EE  \sum_{t=1}^T \left\| \big( \nabla_{\theta} H(\theta_t,\lambda_t) - \sum_{i=1}^N  \lambda_{i,t} G_{i,t}^{\theta}(\hat{\zeta}_{i,t}) \big)\right\|^2 } \nonumber \\
    &\quad +  T D_{\theta} (C_1 \epsilon + C_2 \epsilon^2) \nonumber \\
    &\leq D_{\theta} \sqrt{2T} C_0 +  T D_{\theta} (C_1 \epsilon + C_2 \epsilon^2),
\end{align}
where the third inequality follows from the Jensen's inequality. The fourth inequality follows since 
\begin{align*}
    \EE \left\langle \sum_{i=1}^N \lambda_{i,t} G_{i,t}^{\theta}(\hat{\zeta}_{i,t})] - \nabla_{\theta} H(\theta_t,\lambda_t), \right. \nonumber \\
    \left. \sum_{i=1}^N \lambda_{i,k} G_{i,k}^{\theta}(\hat{\zeta}_{i,k}) - \nabla_{\theta} H(\theta_k,\lambda_k) \right\rangle=0,
\end{align*}
when $t\neq k$. The last inequality follows since 
$\nabla_{\theta} H(\theta_t,\lambda_t) \leq \frac{B_2}{\rho \beta} e^{\frac{B_1}{\rho \beta}}   = C_0$ and  $G_{i,t}^{\theta}(\hat{\zeta}_{i,k} ) \leq C_0$.

Similarly, by virtue of the update rule \eqref{eq:alg:server:lmbda}, for any $\lambda \in \Lambda$, we have
\begin{align}
    &\left\| \lambda_{t+1} - \lambda\right\|^2 \leq \left\| \lambda_t + \eta_{\lambda} g_{t}^{\lambda} - \lambda \right\| \nonumber \\
    & = \left\| \lambda_{t} - \lambda\right\|^2 + \eta_{\lambda}^2 \left\| g_{t}^{\lambda}\right\|^2  + 2 \eta_{\lambda}\langle \lambda_t - \lambda, g_t^{\lambda} \rangle.
\end{align}
It follows that 
\begin{align}\label{eq:centra:term0}
    &\sum_{t=1}^T \langle \lambda - \lambda_t, g_t^{\lambda} \rangle \nonumber \\
    &\leq \frac{1}{2 \eta_{\lambda}} \sum_{t=1}^T \Big( \left\| \lambda_t - \lambda \right\|^2 - \left\| \lambda_{t+1} - \lambda \right\|^2 + \eta_{\lambda}^2 \left\| g_{t}^{\lambda}\right\|^2\Big) \nonumber \\
    &\leq \frac{D_{\lambda}^2}{2\eta_{\lambda}} + \frac{\eta_{\lambda}}{2} \sum_{t=1}^T \left\| g_{t}^{\lambda}\right\|^2 \nonumber \\
    &\leq \frac{D_{\lambda}^2}{2\eta_{\lambda}} + \frac{\eta_{\lambda}  T N C_3^2}{2},
\end{align}
where the last inequality follows since $\left\| g_{i,t}^{\lambda}\right\| \leq e^{\frac{B_1}{\rho \beta}}:= C_3$ and $g_{t}^{\lambda} \leq \sqrt{N} C_3$.

Define $G_{i,t}^{\lambda}(\zeta) = e^{f(\theta_t,\zeta)}/(\rho \beta)$. Note that
\begin{align*}
    \nabla_{\lambda_i} H(\theta_t,\lambda_t) = \EE_{\zeta \sim \hat{\PP}_i} \Big[ e^{f(\theta_t,\zeta)}/(\rho \beta) \Big] = \EE_{\zeta \sim \hat{\PP}_i} [ G_{i,t}^{\lambda}(\zeta) ].
\end{align*}
Again, $g_{i,t}^{\lambda}$ is a biased estimate of the gradient with respect to $\lambda_i$ with the bias error defined by 
\begin{align}
    e_{i,t}^{\lambda} = g_{i,t}^{\lambda} - G_{i,t}^{\lambda}(\hat{\zeta}_{i,t})
\end{align}
Recall that $z_{i,t}^*$ is the unique maximizer of $L(\theta_t,\xi) - \rho c(\xi,\hat{\zeta}_{i,t})$.
It follows that
\begin{align}
    &\left\| e_{i,t}^{\lambda} \right\| \nonumber \\
    &= \Bigg| \exp\left( \frac{L(\theta_t,z_{i,t})- \rho c(z_{i,t},\hat{\zeta}_{i,t}) } {\rho \beta} \right)  \nonumber \\
    & - \exp\left( \frac{L(\theta_t,z_{i,t}^*)- \rho c(z_{i,t}^*,\hat{\zeta}_{i,t}) } {\rho \beta} \right) \Bigg| \nonumber \\
    &\leq e^{\frac{B_1}{\rho \beta}} \Bigg|\frac{L(\theta_t,z_{i,t})- \rho c(z_{i,t},\hat{\zeta}_{i,t}) } {\rho \beta} -  \frac{L(\theta_t,z_{i,t}^*)- \rho c(z_{i,t}^*,\hat{\zeta}_{i,t}) } {\rho \beta} \Bigg| \nonumber \\
    &\leq \frac{1}{\rho \beta} \exp(\frac{B_1}{\rho \beta})L_{\xi} \left\| z_{i,t} - z_{i,t}^*\right\|^2 \nonumber \\
    &\leq \frac{1}{\rho \beta} \exp(\frac{B_1}{\rho \beta})L_{\xi} \epsilon^2 = C_4 \epsilon^2, 
\end{align}
where we define $C_4 = \frac{1}{\rho \beta} \exp(\frac{B_1}{\rho \beta})L_{\xi}$. The first inequality holds since $|e^x - e^y| \leq e^B|x-y|$ when $x,y<B$, the second inequality follows from Assumption~\ref{assump:DRO}.
Then, we have
\begin{align}\label{eq:centra:term1}
    &\max_{\lambda \in \Lambda} \sum_{t=1}^T \left\langle \lambda - \lambda_t ,  \nabla_{\lambda} H(\theta_t,\lambda_t) -g_t^{\lambda}\right\rangle \nonumber \\
    & = \max_{\lambda \in \Lambda} \sum_{t=1}^T \sum_{i=1}^N \left\langle \lambda_i - \lambda_{i,t} ,  \nabla_{\lambda_i} H(\theta_t,\lambda_t) - G_{i,t}^{\lambda}(\hat{\zeta}_{i,t}) \right\rangle \nonumber \\
    &\quad + \sum_{t=1}^T \sum_{i=1}^N \left\langle \lambda_i - \lambda_{i,t}, G_{i,t}^{\lambda}(\hat{\zeta}_{i,t}) - g_{i,t}^{\lambda}  \right\rangle \nonumber \\
    &\leq \max_{\lambda \in \Lambda} \sum_{t=1}^T \sum_{i=1}^N \left\langle \lambda_i - \lambda_{i,t} ,  \nabla_{\lambda_i} H(\theta_t,\lambda_t) - G_{i,t}^{\lambda}(\hat{\zeta}_{i,t}) \right\rangle \nonumber \\
    &\quad + T N C_4 \epsilon^2.
\end{align}
Taking expectation with respect to $\hat{\zeta}_{i,t}$ on both sides of \eqref{eq:centra:term1} yields
\begin{align}\label{eq:centra:term2}
    &\EE \Big[\max_{\lambda \in \Lambda} \sum_{t=1}^T \left\langle \lambda - \lambda_t ,  \nabla_{\lambda} H(\theta_t,\lambda_t) -g_t^{\lambda}\right\rangle \Big]\nonumber \\
    &\leq \EE \Big[ \max_{\lambda \in \Lambda} \sum_{t=1}^T \sum_{i=1}^N \left\langle \lambda_i - \lambda_{i,t} ,  \nabla_{\lambda_i} H(\theta_t,\lambda_t) - G_{i,t}^{\lambda}(\hat{\zeta}_{i,t}) \right\rangle \Big] \nonumber \\
    &\quad + T N C_4 \epsilon^2 \nonumber \\
    &= \EE \Big[ \max_{\lambda \in \Lambda} \sum_{t=1}^T \sum_{i=1}^N \left\langle \lambda_i,  \nabla_{\lambda_i} H(\theta_t,\lambda_t) - G_{i,t}^{\lambda}(\hat{\zeta}_{i,t}) \right\rangle \Big] \nonumber \\
    &\quad + T N C_4 \epsilon^2 \nonumber \\
    &\leq  \EE \Bigg[\sum_{i=1}^N \left\| \sum_{t=1}^T \big( \nabla_{\lambda_i} H(\theta_t,\lambda_t) - G_{i,t}^{\lambda}(\hat{\zeta}_{i,t})\big) \right\| \Bigg] + T N C_4 \epsilon^2 \nonumber \\
    &\leq N \sqrt{2T} C_3 + T N C_4 \epsilon^2,
\end{align}
where the last inequality follows from the Jensen's inequality and the facts that $\nabla_{\lambda_i} H(\theta_t,\lambda_t)\leq e^{\frac{B_1}{\rho \beta}} =  C_3$ and  $G_{i,t}^{\lambda}(\hat{\zeta}_{i,t})\leq C_3$.

Substituting \eqref{eq:centra:t5}, \eqref{eq:centra:t9}, \eqref{eq:centra:term0} and \eqref{eq:centra:term2} into \eqref{eq:centra:t3}, we have
\begin{align}\label{eq:centra:term3}
    & \EE \Big[ \max_{\lambda \in \Lambda} H(\bar{\theta}, \lambda) - \min_{\theta \in \Theta} \max_{\lambda \in \Lambda} H(\theta, \lambda) \Big]  \nonumber \\
    &\leq \frac{D_{\theta}^2}{2\eta_{\theta} T} + \frac{\eta_{\theta}  C_0^2}{2} + \frac{D_{\theta} \sqrt{2} C_0}{\sqrt{T}} +   D_{\theta} (C_1 \epsilon + C_2 \epsilon^2) \nonumber \\
    &\quad + \frac{D_{\lambda}^2}{2\eta_{\lambda} T} + \frac{\eta_{\lambda}  N C_3^2}{2} + \frac{N \sqrt{2} C_3}{\sqrt{T}} +  N C_4 \epsilon^2.
\end{align}
Substituting $\eta_{\lambda} = \eta_{\theta} =\frac{1}{\sqrt{T}} $ into \eqref{eq:centra:term3} completes the proof.

\bibliography{autosam}
\bibliographystyle{unsrt} 

\end{document}